\documentclass{llncs}

\usepackage[utf8]{inputenc}
\usepackage[T1]{fontenc}
\usepackage[american]{babel}

\usepackage{tikz}
\usetikzlibrary{shapes,arrows,automata,plotmarks,positioning}

\usepackage{subcaption}
\usepackage{calc}
\usepackage{amssymb}
\usepackage{amstext}
\usepackage{amsmath}
\usepackage{amsthm}
\usepackage{pslatex}

\usepackage{etoolbox,siunitx}
\robustify\bfseries
\usepackage{booktabs}
\usepackage{multirow}
\usepackage{hyperref}
\usepackage[inline]{enumitem}
\usepackage[capitalise,noabbrev]{cleveref}
\usepackage{float}

\newcommand{\wue}{Würzburg}
\newcommand{\shw}{Bad Schwartau}
\newcommand{\ind}{Markt Indersdorf}
\newcommand{\hobos}{HOBOS}
\newcommand{\webee}{we4bee}
\newcommand{\jel}{Jelgava}
\newcommand{\algo}{RBA}

\newcommand{\wueshort}{Wü.}
\newcommand{\shwshort}{B. S.}
\newcommand{\indshort}{M. I.}
\newcommand{\jelshort}{Jel.}

\newcommand{\overallshort}{All}

\newcommand{\para}[1]{\smallskip\noindent\textbf{#1}.}

\title{Anomaly Detection in Beehives: An Algorithm Comparison}

\author{
Padraig Davidson\inst{1}
\and Michael Steininger\inst{1}
\and Florian Lautenschlager\inst{1}
\and Anna Krause\inst{1}
\and Andreas Hotho\inst{1}
}

\institute{
Institute of Computer Science, Chair of Computer Science X, University of Würzburg, Am Hubland, Würzburg, Germany,\\
\email{\{davidson, steininger, lautenschlager, anna.krause, hotho\}@informatik.uni-wuerzburg.de}
}

\begin{document}
\maketitle

\begin{abstract}
Sensor-equipped beehives allow monitoring the living conditions of bees.
Machine learning models can use the data of such hives to learn behavioral patterns and find anomalous events.
One type of event that is of particular interest to apiarists for economical reasons is bee swarming.
Other events of interest are behavioral anomalies from illness and technical anomalies, e.g. sensor failure.
Beekeepers can be supported by suitable machine learning models which can detect these events.

In this paper we compare multiple machine learning models for anomaly detection and evaluate them for their applicability in the context of beehives.
Namely we employed Deep Recurrent Autoencoder, Elliptic Envelope, Isolation Forest, Local Outlier Factor and One-Class SVM.
Through evaluation with real world datasets of different hives and with different sensor setups we find that the autoencoder is the best multi-purpose anomaly detector in comparison.

\end{abstract}

\keywords{
Precision Beekeeping
\and Anomaly Detection
\and Deep Learning
\and Autoencoder
\and Swarming
}

\section{Introduction}
\label{sec:introduction}
Supporting beekeepers in their care decisions is the goal of precision apiculture.
To this end, sensors are used which collect data on
\begin{itemize*}
    \item[1)] apiary-level (i.e. meteorological parameters),
    \item[2)] colony-level (i.e. beehive temperature), or
    \item[3)] individual bee-related level (i.e. bee counter)
\end{itemize*}~\cite{zacepins2015challenges}.
For colony-level data, environmental sensors are installed in beehives in order to monitor and quantify the beehive's state continuously.
Sensor values we expect most of the times are defined as normal regions of observations, while values differing considerably from this norm are called anomalies.
Defining norm and anomaly is always contingent on the context of the analysis.
We differentiate between behavioral anomalies, sensor anomalies and external interference.
The first anomaly type is characterized by irregular behavior of the bees, the second type occurs when there are irregular measurements due to the sensors, and the last type represents anomalies induced by any external force.

An important behavioral anomaly for beekeepers is swarming, which describes a queen leaving her hive accompanied by worker bees in order to establish a new colony.
First, there is the \textit{prime swarm} where the current queen leaves the hive with a large number of worker bees.
This can be followed by multiple \textit{after swarms} with fewer workers departing.
These events can even lead to the complete depletion of a colony~\cite{winston1980swarming}.
Beekeepers want to prevent swarming as it reduces honey production.
Additionally, swarming requires immediate action to recollect the new colonies.
Due to the highly stochastic nature of this reproduction process the prediction of these events is difficult.

anomalies which are not directly related to bees can also occur.
On the one hand, there are sensor anomalies which are caused by defective sensors.
These require repair in order to restore a beehive's complete functionality.
On the other hand, there can be anomalies due to external interference.
This usually occurs through physical interaction of the beekeeper with the hive, e.g. when the hive is opened to yield honey.


For large datasets of beehive data it is infeasible to find anomalies manually.
Therefore, we apply automatic anomaly detection methods.
A number of machine learning algorithms have shown to provide this functionality in other domains.
It is therefore interesting to assess how these methods perform in the context of beekeeping.


In this work, we evaluate multiple common anomaly detection models, namely Deep Recurrent Autoencoders, Elliptic Envelope, Isolation Forests, Local Outlier Factor and One-Class SVMs, for their applicability with beehive data.
We evaluate these models on three datasets for this work:
Two short term datasets, one from~\cite{zacepins2016remote} and the other from \webee\ (\url{https://we4bee.org/}), and one long term dataset from the \hobos\ (\url{https://hobos.de/}) project containing four years of data.
These datasets contain labelled swarming events (e.g. observed events by the apiarist) and other anomalies without labels (e.g. hidden or unobserved).
The models are trained to find anomalies based on temperature readings of a beehive in an univariate setting (with one temperature sensor) and in a multivariate setting (with three temperature sensors).
We use the labelled swarms to assess anomaly detection performance of our models quantitatively.
Our results suggest that recurrent autoencoders provide consistently good results across the datasets for both, the univariate and the multivariate setting, compared to the other models.
Elliptic Envelope's performance is inconsistent, since it showed by far the best performance when trained on one beehive but also the worst when trained on another beehive.
This implies, prediction quality is strongly dependent on the training data.
The other models have also shown to provide relatively good performance.
Furthermore, we present other types of anomalies found through automatic anomaly detection, namely through the recurrent autoencoder, for which no labels exist and discuss the usage of anomaly detection for non-swarming anomalies.

Our contribution is twofold:
First, we compare typical anomaly detection machine learning models for swarm detection in both a univariate and a multivariate sensor setting.
Second, we present other types of anomalies found by the recurrent autoencoder in the beehive datasets and discuss anomaly detection for these anomalies.

In this work we present an extension of our work in \cite{davidson2020anomaly}.
This includes a broader spectrum of anomaly detectors, not solely the recurrent autoencoder.
Furthermore we added a quantitative analysis of the swarm prediction quality of all detectors.
The analysis was done in a univariate sensor setting, as well as a multivariate setting.

This work is structured as follows:
Related research is presented in \cref{sec:rel_work}.
\cref{sec:dataset} describes the datasets used in this work.
The different anomaly detection models of our comparison are presented in \cref{sec:methods}.
A description of our experiments can be found in \cref{sec:experiments} while their results are shown in \cref{sec:results}.
We discuss our results in \cref{sec:discussion} before concluding the work in \cref{sec:conclusion}.

\section{Related Work}
\label{sec:rel_work}
\noindent
There are a number of works which encompass monitoring and detection of swarms in beehives. 

Ferrari et al.~\cite{ferrari2008monitoring} analyzed humidity, temperature and sound in beehives to understand how these variables change before and during swarming.
To this end, they used data from three beehives where nine swarming events had occurred.
The authors identified that a change in temperature and a shift in sound frequency might be useful indicators for swarming.


Kridi et al.~\cite{kridi2014predictive} determined pre-swarming behavior through clustering temperature data.
If measurements cannot be assigned to clusters of typical beehive temperature patterns for several hours, the authors consider this an anomaly. 


Zacepins et al.~\cite{zacepins2016remote} proposed an rule-based algorithm for swarming detection using data from a single temperature sensor.
Their algorithm (from here on denoted as \algo) detects a swarming event if the temperature is above \SI{35.5}{\celsius} for between two and twenty minutes.
Events with shorter or longer temperature anomalies are not considered to be swarms.


Zhu et al.~\cite{zhu2019increase} link a linear rise in temperature to pre-swarming behavior.
They recommend placing a temperature sensor between the bottom of the first frame and the beehive's wall, as this is the most suited location for measuring this increase in temperature.



While some of these works propose swarm detection methods, none of them evaluated a larger set of common machine learning approaches for anomaly detection.
Popular models include One-Class SVMs~\cite{scholkopf2001estimating}, Local Outlier Factor (LOF)~\cite{breunig2000lof}, Elliptic Envelope~\cite{rousseeuw1999fast}, Isolation Forests~\cite{liu2008isolation} and neural networks~\cite{ryan1998intrusion}.
As for neural networks, recurrent autoencoders performed particularly well on sequential data across many anomaly detection settings~\cite{filonov2016multivariate,malhotra2016multi,shipmon2017time,chalapathy2019deepad}.
Therefore, we evaluate these algorithms to identify which is the most promising for this task.

\section{Datasets}
\label{sec:dataset}
\noindent
We obtained datasets from three sources for our studies: \hobos, \webee, and a subset of Zacepins et al.~\cite{zacepins2016remote} dataset.
We selected two \hobos\ beehives in Würzburg and Bad Schwartau and one \webee\ hive in \ind\  for our experiments. 
Zacepins et al. data was collected in \jel.
From here on, we refer to all datasets by the location of the beehive.

\subsection{\wue\ \& \shw}
\begin{figure}[b!]
    \centering
    \includegraphics[width=0.7\linewidth]{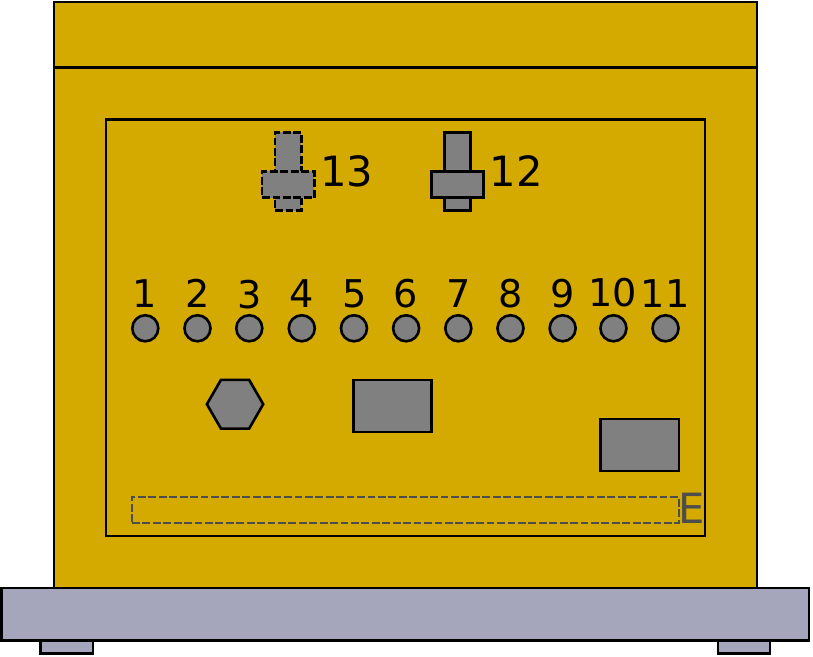}
    \caption{Back of a \hobos\ beehive.
    Temperature sensors T$_1$--T$_{11}$ are mounted between honeycombs, temperature sensors T$_{12}$ and T$_{13}$ are mounted on the back and the front of the hive, respectively.
    E denotes the hive's entrance on the front of the beehive.\cite{davidson2020anomaly}}
    \label{fig:hobos_sensors}
\end{figure}

\hobos\ collected evironmental data from five sensor equipped beehives (species \textit{apis mellifera}; beehive type: zander beehive) in Bournemouth, Münchsmünster, Gut Dietlhofen, \shw\ and \wue.
We selected the hives in \shw\ as there are three verified swarming events.
In contrast, data for the \wue\ beehive is completely unverified.
We use this beehive to assess cross-beehive applicability of our models.
\hobos\ beehives come with different sensor configurations.
\cref{fig:hobos_sensors} shows the maximum sensor configuration: 13 temperature sensors, named T$_1$ to T$_{11}$ mounted between the honeycombs and T$_{12}$ and T$_{13}$ mounted on the back and front of the hive respectively, plus weight, humidity, and carbon dioxide (CO$_2$) sensors.
The beehives in \shw\ and \wue\ are both missing some of the temperature sensors:
\shw\ is not equipped with T$_2$, T$_3$, T$_9$, T$_{10}$ and T$_{12}$ and \wue\ is not equipped with T$_{2}$ and T$_{3}$.
\hobos\ collected data from May 2016 to September 2019.
During this time, sensor readings were collected once per minute for every sensor.
As the typical swarming period for honey bees is May to September \cite{fell1977seasonal}, we limit data for our preliminary study to the swarming period.
\hobos\ granted us access to their complete dataset.

\para{Analysis}
The Pearson correlation coefficients between different sensors, e.g. inter-sensor correlations, are visualized in \cref{fig:correlation_anomalies} in the appendix.
Within the normal data portion of the dataset, these correlations are strong, especially between adjacent sensors.
Correlations are even higher for the sensors T$_4$--T$_{10}$ placed in the center of the beehive, and go beyond directly adjacent sensors, i.e. sensors T$_4$ and T$_{10}$ still correlate positively.
The sensors placed at the outer margins of the apiary tend to correlate with the sensors placed outside, as well as their opposite counterpart.

During days containing anomalies, correlations are not that strong, except for neighboring sensors.
This implies, that certain sensors are more sensible for swarm detection, as also stated in~\cite{zhu2019increase}.


\subsection{\jel}
Zacepins et al. monitored ten colonies (\textit{apis mellifera mellifera}; norwegian-type hive bodies) with a single temperature sensor placed above the hive.
The observation ran from May to August in 2015 and recorded one measurement per minute.
The authors recorded nine swarming events during their observation period and granted us access to the nine days in the dataset that contain these events.


\subsection{\ind}
\begin{figure}[b!]
    \centering
    \includegraphics[width=0.7\linewidth]{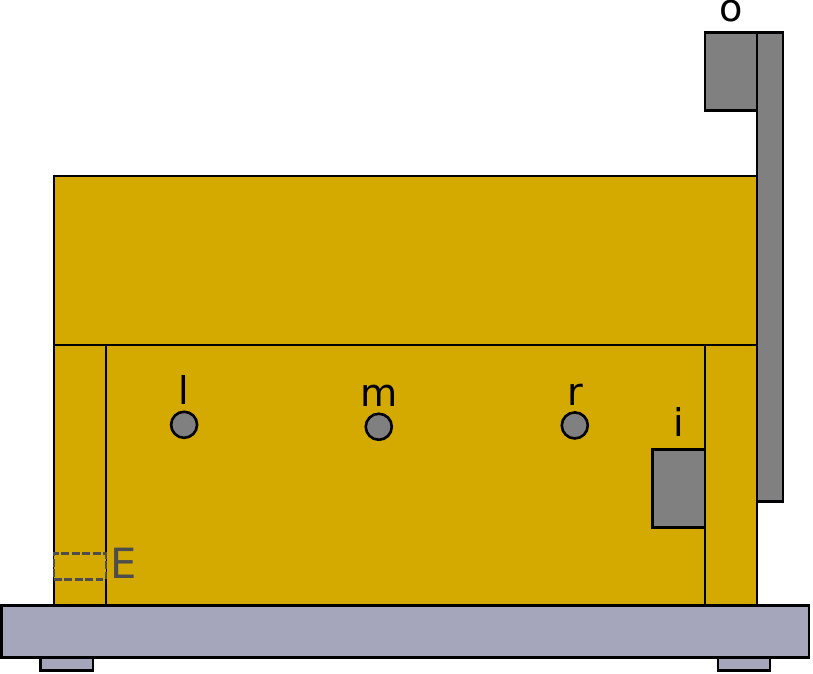}
    \caption{Cutaway view of a \webee\ beehive.
    T$_l$, T$_m$, T$_r$, and T$_i$ are mounted on the inside, laterally to the honeycombs.
    T$_o$ is placed outside at the pylon.
    E denotes the entrance on the front of the beehive.\cite{davidson2020anomaly}}
    \label{fig:we4bee_sensors}
\end{figure}

\webee\ started rolling out 100 smart top bar hives to schools and interested individuals all over Germany in 2019.
In the same year, first bee colonies have been introduced into the hives.
One successful hive of this first project year is the hive in \ind\ (\textit{apis mellifera}; top bar hive).
\cref{fig:we4bee_sensors} shows the cutaway view of a \webee\ hive: it includes four temperature sensors on the inside of the hive and one on the outside.
Three temperature sensors inside the hive distributed along the length of the hive, the fourth is located at the back.
The inner temperature sensors are referred to as T$_l$, T$_m$, and T$_r$ for the sensors in the hive body; the sensor at the back is named T$_i$.
The outside sensor is called T$_o$.
\webee\ hives also report other environmental quantities: air pressure, weight, fine dust, humidity, rain and wind.
For \ind\ we obtained data from June (when the colony was introduced to the hive) to September 2019.
All sensors except fine dust reported one measurement per second; fine dust was recorded once every three minutes.


\section{Methods}
\label{sec:methods}
\subsection{(Recurrent) Autoencoder}
\label{sec:autoencoder}
An autoencoder (AE) consists of two neural networks, an encoder $\phi$ and a decoder $\psi$.
The encoder maps the input space $\mathcal{X}$ into the feature space ($\phi: \mathcal{X} \rightarrow \mathcal{F}$).
In contrast, the decoder remaps the feature space into the input space ($\psi: \mathcal{F} \rightarrow \mathcal{X}$).
The task of the encoder-decoder pair is to adapt both mapping steps in a way, that decoding an encoded sequence closely resembles the input itself: $\bar{x} = \psi(\phi(x)) \sim x$.
When training the AE with normal data, this kind of data is encoded very well within the feature space, whereas anomalous data cannot be reconstructed properly, incurring a high difference in prediction and input.

This difference is quantified by a loss function $\mathcal{L}$, which is often the $l_2$ norm~\cite{zhou2017anomaly} or the MSE (mean squared error)~\cite{shipmon2017time}.
The optimization task can be stated as follows:
\[ \phi, \psi = \text{arg min}_{\phi, \psi} \mathcal{L}(x, \psi(\phi(x))). \]
An anomaly is any input with a resulting loss of greater than $\alpha$, which is the anomaly threshold: $\mathcal{L}(x, \bar{x}) \geq \alpha$.
This hyperparameter can either be set manually or determined with a labelled anomaly set in a second training step.
Optimally, $\alpha$ is set high enough to detect all anomalies, but no too low to be overly sensible within predictions[\textbf{Find cite}].

\subsection{Local Outlier Factor}
The local outlier factor \cite{breunig2000lof} estimates the degree or probability of an instance being an anomaly, rather than performing a binary classification.
The algorithm is based on the idea of density based clustering, which generally requires two parameters: a minimum number of objects $k$ and a volume value.
Together, these parameters define a local density.
Regions with densities higher than the density threshold form clusters and are separated by regions with densities below the density threshold.

As an extension of this idea, the local outlier factor algorithm only relies on one parameter, the minimum number of neighbors $k$.
Densities are calculated by using the $k$-heighborhood.
The local reachability density is the average reachability distance of a point to its $k$ neighborhood.
The reachability distance of two points is the maximum of the $k$-distance or the distance between the two points: $rd_k(A,B) := \max \{k-\text{distance}(B), d(A,B)\}$ \cite{breunig2000lof}.
Finally, the local outlier factor of a point is calculated as the average local reachability density of the $k$ neighborhood divided by the local reachability of a given point.

A local outlier factor of $\leq 1$ indicates an (cluster) inlier, whereas values $>1$ indicate outliers.

\subsection{Isolation Forest}


While most anomaly detectors build internal profiles of normal data and report anomalies that do not fit in these profiles, the isolation forest is based on the idea of isolating anomalies \cite{liu2008isolation,liu2012isolation}.
An isolation forest consists of several isolation trees, or iTrees.
An iTree is a binary search tree, which consists of external nodes (e.g. nodes without children) and internal nodes (e.g. nodes with exactly two children).
Internal nodes split the data by a random attribute $q$ and a corresponding split value $p$. 
If a data point fullfills the ``test'' function $q < p$ the path to the first child is followed, otherwise the second path is pursued.
This structure is repeated until all data points $x$ in the dataset are isolated in an external node.

Since anomalies are isolated more easily, the path length $h(x)$ for an anomalous point is shorter than for normal data.
An anomaly score can be computed as $s(x,m) = \text{pow}(2, -E(h(x))/c(m)$, where $E(h(x))$ is the average $h(x)$ from all iTrees and $c(m)$ represents the average path length given the sample size $m$.
The anomaly score indicates an anomaly if it is close to $1$ and normal data for values close to or below $0.5$.

\subsection{Elliptic Envelope}
Elliptic envelope is an anomaly detection algorithm that is based on the minimum covariance determinant (MCD) estimator and assumes the data to be sampled from an elliptically symmetric unimodal distribution.
MCD is a highly robust estimators of multivariate location and scatter~\cite{rousseeuw1984least}.

The method subsamples the data $\mathbf{X}$ in $\mathbf{H}_1$ and computes an estimate of the location $\mathbf{T}_1$ and the covariance of each sample $\mathbf{S}_1$.
A new subsample $\mathbf{H}_2$ is built with the $h = (n + p + 1)/2$ samples with the lowest robust distance~\cite{rousseeuw1999fast}, where $n$ is the number of samples in $\mathbf{H}_1$ and $p$ the number of features.
This subsampling process is repeated until the determinant of the covariance converges within a given tolerance.
Elliptic envelope finally flags every sample as outlier that has a robust distance above a cutoff value $\sqrt{\chi^2_{p,0.975}}$~\cite{rousseeuw1999fast}.

\subsection{One-Class SVM}
The one-class support vector machine (SVM)~\cite{li2003improving}, is an extension of the standard support vector machine~\cite{cortes1995svm} for unsupervised outlier detection.
The SVM algorithm is normally applied to supervised two class classification problems.
Input is classified by finding a hyperplane with maximum distance to the closest instances of the two classes.
Data points of the same class, in our setting normal data or anomalous data, are grouped on the same side of this plane.
To account for datapoints still being non-seperable, a penalty parameter is introduced.
The One-Class SVM reuses the the SVM algorithm by setting all class labels to the same class.
This means, the separating hyperplane is an envelope around the normal data, with a maximum distance towards all anomalies.

\section{Experimental Setup}
\label{sec:experiments}
All models described in~\cref{sec:methods} are evaluated on the datasets outlined in~\cref{sec:dataset}.

\para{Data splitting}
We used the \hobos\ hives for training and validation purposes.
That is, we trained on \shw\ and \wue\ in independent settings using the reported and found \cite{davidson2020anomaly} swarming events.
Explicitly, we built two setups: one with training the models on the normal behavior of \shw, using its anomalous behavior as a validation set for the parameter search, and one with the training step consisting of the normal behavior of \wue, while validating on its anomalous behavior set.
The datasets from \jel\ and \ind, as well as the test set from the untrained hive were only used for evaluation.
All models were provided with the same splits of input data to ensure comparability.

\begin{figure}[b!]
        \centering
        \includegraphics[width=0.7\textwidth]{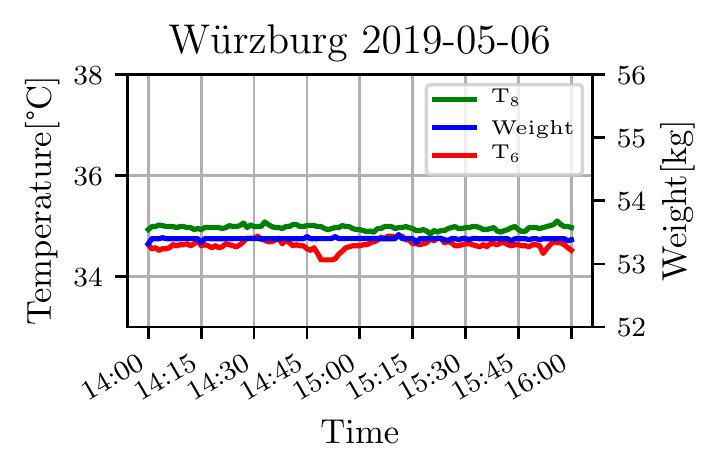}
        \caption{Normal behavior of all three sensors. \cite{davidson2020anomaly}}
        \label{fig:colony_behavior_normal}
\end{figure}

As customary in novelty detection (i.e. AE, Isolation Forest), the training data shouldn't be polluted by outliers.
For that, we visually examined the datastream and marked each day as normal behavior, or as an outlier, e.g. an anomaly.
We defined normal behavior as any temperature sensor trace remaining nearly constant at \SI{34.5}{\celsius} as the core temperature \cite{ferrari2008monitoring,zacepins2016remote}.
Any larger deviation form this norm temperature were considered as anomalous days.
\cref{fig:colony_behavior_normal} shows sensor data to be expected from a normal day.
Training and validation parts consist exclusively of normal data, while test and holdout sets combine normal and anomalous data.
Test sets are any portions of the dataset with anomalous data, that don't originate from the beehive used for training.
The holdout set, which contains the anomalous behavior from the training hive, is used for the parameter search of the estimators.

An exemplary view of the data splitting procedure for the \shw\ hive can be seen in~\cref{fig:data_splitting}.
\begin{figure}[b!]
    \centering
    \resizebox{.7\linewidth}{!}{
        \begin{tikzpicture}
            \usetikzlibrary{calc}
            \usetikzlibrary{decorations.pathreplacing}
            \tikzstyle{block}=[rectangle,draw,align=center,minimum height=2em, outer sep=0cm]
            \tikzstyle{train}=[text width=7.5em]
            \tikzstyle{val}=[text width=4.5em]
            \tikzstyle{test}=[text width=4.5em]
            
            \tikzstyle{nameblock}=[rectangle,align=right,text width=7em,minimum height=2em]
            \tikzstyle{labelblock}=[rectangle,align=center,text width=7em,minimum height=2em]
            \tikzstyle{greycolor}=[draw=black!25, color=black!25]
        
            \node[block, test] (WueAnomaly) {Holdout\\\small{\SI{69132}{\minute}}};
            \node[block, val, left=0.2cm of WueAnomaly] (WueVal) {Validation\\\small{\SI{75314}{\minute}}};
            \node[block, train, left=0cm of WueVal] (WueTrain) {Training\\\small{\SI{677822}{\minute}}};
            \node[nameblock, left=0.2cm of WueTrain] (Wue) {\textbf{\shw}};
            
            \node[block, test, below=0.5cm of WueAnomaly] (BadSAnomaly) {Test\\\small{\SI{72013}{\minute}}};
            \node[block, val, greycolor, left=0.2cm of BadSAnomaly] (BadSVal) {Validation\\\small{\SI{46082}{\minute}}};
            \node[block, train, greycolor, left=0cm of BadSVal] (BadSTrain) {Training\\\small{\SI{42134}{\minute}}};
            \node[nameblock, left=0.2cm of BadSTrain] (BadS) {\wue};
            
            \node[block, test, below=0.5cm of BadSAnomaly] (JelAnomaly) {Test\\\small{\SI{12960}{\minute}}};
            \node[nameblock, below=0.5cm of BadS] (Jel) {\jel};
            
            \node[block, test, below=0.5cm of JelAnomaly] (IndAnomaly) {Test\\\small{\SI{100021}{\minute}}};
            \node[nameblock, below=0.5cm of Jel] (Ind) {\ind};
            
            \draw [decorate,decoration={brace,amplitude=10pt, raise=0.1cm}] (WueTrain.north west) -- (WueVal.north east) node [midway,above=0.6cm] {Normal Behavior};
            \draw [decorate,decoration={brace,amplitude=10pt, raise=0.1cm}] (WueAnomaly.north west) -- (WueAnomaly.north east) node [midway,above=0.6cm] {Anomalous Behavior};
            
        \end{tikzpicture}
    }
    
    \caption{The data splits used for \shw. The autoencoder is trained on \shw's `Training'.
    The hyperparameters and \(\alpha\) are tuned using its `Validation' and `Holdout', respectively.
    The model is then tested on all `Test'.
    For \wue, the splits are set accordingly using its `Training', `Validation', and `Test' as `Holdout'.
    We provide the recording time for all splits.~\cite{davidson2020anomaly}
    }
    \label{fig:data_splitting}
\end{figure}
Keep in mind, that the test and holdout sets also contain slices of normal data and are not necessarily only windows with anomalies.

\para{Input data}
In any univariate sensor setting, we use central temperature sensors.
For the hives in \wue\ and \shw, those are T$_6$ -- T$_8$, from which we evaluate data on T$_6$ and T$_8$.
In \ind\ this is T$_m$, which we additionally downsampled to one minute resolution to be consistent with the other datasets.
For \jel\ the single temperature sensor at the top is used.
In the multivariate sensor setting, we used senors T$_6$, T$_7$, T$_8$ for the hives \shw\ and \wue, whereas we used T$_l$, T$_m$, T$_r$ in \ind.

Any model is given a \SI{60}{\minute} window of sensor data, which corresponds to 60 consecutive input values of temperature data per sensor.
According to \cite{zhu2019increase,zacepins2016remote,ferrari2008monitoring} swarming events last from \SI{20}{\minute} to \SI{60}{\minute} in duration.

In the multivariate sensor setting, the AE is provided with sensor data of $60\times3$, whereas the other models are given $60\cdot3$ values, i.e. concatenating the three sensors.

Input data for the AE was normalized via standard scaling (e.g. their z-score).
The non-autoencoder models were provided with the raw and unscaled sensor values, since scaling impaired their predictions.

\para{Model training}
The optimal parameter settings for the models were found employing a random search \cite{bergstra2012random}.
A table with all parameters that were optimized is given in the appendix (see \cref{tab:appendix_hyperparameters}).
Parameters were optimized using the normal data of a beehive, while using the anomalous behavior of that hive as the validation set for this search.
The $F_1$ score of predicting swarms was used as the metric to be maximized.

For the models Local Outlier Factor, Elliptic Envelope, Isolation Forest and One-Class SVM we relied on the implementations in \cite{scikit-learn2011sciki-learn}.
In the same setting we searched for the remaining parameter $\alpha$ for the pre-trained AE, which we could not do in \cite{davidson2020anomaly} due to missing labels.
Within this (second) parameter setting step for the swarm detection, we used a windowing technique, shifting the window by 15 minutes forward in time to extract the next window.

Pre-training of the AE was done in a preliminary random search (see appendix), finding the best hyperparameters for the reconstruction task itself.
The \textit{Adam} optimizer \cite{kingma2014adam} was used with the default parameters ($lr=10^{-3}$) and the mean squared error (MSE) as the loss function.
Early stopping with five epoch patience was employed
to prevent overfitting.
For pre-training, we used all suitable measurement windows by shifting the window one time step further.

\para{Predictions}
\algo\ \cite{zacepins2016remote} is utilized on all anomalous behavior subsets to predict swarming events.
We additionally used it on the normal behavior portions of the dataset to ensure no swarming events in the training steps of both training hives.

Predictions with all other models were made by using the best model configuration found in the random search, while predicting events within the tests sets, e.g. the anomalous sets.
For instance, a model was trained on \shw, using its holdout set for the grid search, while predicting the anomaly sets of \wue, \jel\ and \ind.

\para{Evaluation}
To evaluate the various classifiers, we used standard classification metrics.
\textit{True positives} (TP) contain all time series correctly classified as swarms, whereas \textit{true negatives} (TN) represent all time series correctly labelled as non-swarms.
The other two units quantify miss-classifications.
\textit{false positives} (FP) are any non-swarms classified as swarms, and \textit{false negative} (FN) any swarms categorized as non-swarms.

With these quantifiers, we can calculate performance measures of the classifiers:
\[ P := \frac{TP}{TP + FP} \quad R := \frac{TP}{TP + FN} \quad F_1 := \frac{2 \cdot P \cdot R}{P + R}\]
where $P$ represents the precision, $R$ the recall and $F_1$ the $F_1$-measure.

\section{Results}
\label{sec:results}
\subsection{Univariate}
\cref{tab:results_univariate}
lists the classification metrics for swarming events in the univariate sensor setting for temperature sensor T$_8$ on the hives \wue\ and \shw\ respectively.
The left hand side shows classification metrics using \shw\ as the training hive, the right hand side shows this for \wue.
The best results in the category precision and $F_1$ are highlighted in bold, except for \algo.
For full disclosure, \ind\ is listed in this table, too, but since there are no true positives for swarms, the metrics are degrade and therefore it is not taken into account for calculations.
\begin{table}[t!]
    \centering
    \caption{Overview of classification metrics and results.
    Results are only calculated by the true positives of swarms!
    The estimators are trained on \shw\ on the left hand side and \wue\ on the right hand side (separated by vertical double lines), both with sensor T$_8$.
    Precision (P), Recall (R), and $F_1$ are reported, and set to 0 for no correct classification and $F_1$ set as NA.
    Corresponding true positives (TP), false positives (FP), false negatives (FN), and true negatives (TN) values are also reported.
    The overall metrics are calculated from the weighted scores from each hive.
    }
    \label{tab:results_univariate}
    \scalebox{0.95}{
    \begin{tabular}{@{}ll|S[round-mode=places,round-precision=2]S[round-mode=places,round-precision=2]S[round-mode=places,round-precision=2]|cccc||l|S[round-mode=places,round-precision=2]S[round-mode=places,round-precision=2]S[round-mode=places,round-precision=2]|cccc@{}}
        \toprule
        Classifier & Hive & {P[\%]} & {R[\%]} & {$F_1$[\%]} & TP & FP & TN & FN & Hive & {P[\%]} & {R[\%]} & {$F_1$[\%]} & TP & FP & TN & FN\\
        \midrule
        \multirow{3}{*}{\parbox{1cm}{Local\\Outlier\\Factor}} & \jelshort & 0.3157894736842105 & 1.0 & 0.4799999999999999 & 36 & 78 & 723 & 0 & \jelshort & 0.06923076923076923 & 1.0 & 0.12949640287769784 & 36 & 484 & 317 & 0\\
        & \wueshort & 0.01020408163265306 & 0.625 & 0.02008032128514056 & 5 & 485 & 4268 & 3 & \shwshort & 0.00607748797163839 & 1.0 & 0.012081550465643092 & 24 & 3925 & 716 & 0\\
         & \overallshort & 0.05589450198762869 & 0.6810691318327974 & 0.08884644688077066 & 41 & 563 & 4991 & 3 & \overallshort & 0.015684775578670835 & 1.0 & 0.029943460947084354 & 60 & 4409 & 1033 & 0\\
        \cmidrule{2-17}
        & \indshort & 0.0 & 0.0 & {NA} & 0 & 1532 & 5185 & 0 & \indshort & 0.0 & 0.0 & {NA} & 0 & 303 & 6414 & 0\\
        
        \midrule
        
        \multirow{3}{*}{\parbox{1cm}{Elliptic\\Envelope}} & \jelshort & 0.5 & 0.9722222222222222 & 0.660377358490566 & 35 & 35 & 766 & 1 & \jelshort & 0.17142857142857143 & 1.0 & 0.2926829268292683 & 36 & 174 & 627 & 0\\
        & \wueshort & 0.0 & 0.0 & {NA} & 0 & 51 & 4702 & 8 & \shwshort & 0.005536514632217242 & 0.875 & 0.01100340581608593 & 21 & 3772 & 869 & 3\\
        & \overallshort & 0.0747588424437299 & 0.14536441586280816 & 0.09873809379360554 & 35 & 86 & 5468 & 9 & \overallshort & 0.030773092519994133 & 0.8940158124318429 & 0.05385432531591029 & 57 & 3946 & 1496 & 3\\
        \cmidrule{2-17}
        & \indshort & 0.0 & 0.0 & {NA} & 0 & 58 & 6659 & 0 & \indshort & 0.0 & 0.0 & {NA} & 0 & 17 & 6700 & 0\\
        
        \midrule
        
        \multirow{3}{*}{\parbox{1cm}{Isolation\\Forest}} & \jelshort & 0.3253012048192771 & 0.75 & 0.45378151260504196 & 27 & 56 & 745 & 9 & \jelshort & 0.20915032679738563 & 0.8888888888888888 & 0.33862433862433866 & 32 & 121 & 680 & 4\\
        & \wueshort & 0.0016346546791990192 & 0.5 & 0.003258655804480652 & 4 & 2443 & 2310 & 4 & \shwshort & 0.0016346546791990192 & 0.5 & 0.003258655804480652 & 4 & 2443 & 2310 & 4\\
        & \overallshort & 0.05002852793165443 & 0.537379421221865 & 0.07061979034218516 & 31 & 2499 & 3055 & 13 & \overallshort & 0.03266191755217547 & 0.5581457663451233 & 0.05340175629040797 & 36 & 2564 & 2990 & 8\\
        \cmidrule{2-17}
        & \indshort & 0.0 & 0.0 & {NA} & 0 & 5226 & 1491 & 0 & \indshort & 0.0 & 0.0 & {NA} & 0 & 4056 & 2661 & 0\\
        
        \midrule
        
        \multirow{3}{*}{\parbox{1cm}{One-Class\\SVM}} & \jelshort & 0.5882352941176471 & 0.8333333333333334 & 0.6896551724137931 & 30 & 21 & 780 & 6 & \jelshort & 0.29906542056074764 & 0.8888888888888888 & 0.44755244755244755 & 32 & 75 & 726 & 4\\
        & \wueshort & 0.0 & 0.0 & {NA} & 0 & 298 & 4455 & 8 & \shwshort & 0.00922671353251318 & 0.875 & 0.01826086956521739 & 21 & 2255 & 2386 & 3\\
        & \overallshort & \bfseries 0.08795157934556459 & 0.12459807073954984 & 0.10311564474997229 & 30 & 319 & 5235 & 14 & \overallshort & 0.05331886143920752 & 0.8771128680479825 & 0.08356749456981784 & 53 & 2330 & 3112 & 7\\
        \cmidrule{2-17}
        & \indshort & 0.0 & 0.0 & {NA} & 0 & 1363 & 5354 & 0 & \indshort & 0.0 & 0.0 & {NA} & 0 & 204 & 6513 & 0\\
        
        \midrule
        
        \multirow{4}{*}{AE} & \jelshort & 0.5692307692307692 & 1.0 & 0.7254901960784313 & 37 & 28 & 772 & 0 & \jelshort & 0.5 & 1.0 & 0.6666666666666666 & 36 & 36 & 765 & 0\\
        & \wueshort & 0.00784313725490196 & 0.5 & 0.015444015444015444 & 4 & 506 & 4247 & 4 & \shwshort&0.008936170212765958&0.875&0.017691659646166806& 21 & 2329 & 2312 & 3\\
        & \overallshort & \bfseries 0.09178051631238694 & 0.5747588424437299 & \bfseries 0.12160847653565642 & 40 & 535 & 5019 & 4 & \overallshort & \bfseries 0.08363999164713799 & 0.8940158124318429 & \bfseries 0.11641795569781319 & 57 & 2365 & 3077 & 3\\
        \cmidrule{2-17}
        & \indshort & 0.0 & 0.0 & {NA} & 0 & 251 & 6466 & 0 & \indshort & 0.0 & 0.0 & {NA} & 0 & 1934 & 4783 & 0\\

        \midrule
        \midrule
        
        \multirow{3}{*}{\algo} & \jelshort & 1.0 & 0.5 & 0.6666666666666666 & 18 & 0 & 801 & 18 & \jelshort & 1.0 & 0.5 & 0.6666666666666666 & 18 & 0 & 801 & 18\\
        & \wueshort & 0.06896551724137931 & 0.25 & 0.1081081081081081 & 2 & 27 & 4726 & 6 & \shwshort & 0.5714285714285714 & 0.3333333333333333 & 0.4210526315789474 & 8 & 6 & 4635 & 16\\
        & \overallshort & 0.2081716376538419 & 0.28737942122186494 & 0.19162249065786044 & 20 & 27 & 5527 & 24 & \overallshort & 0.6366256426234617 & 0.35868774990912394 & 0.4584170349537967 & 26 & 6 & 5436 & 34\\
        \cmidrule{2-17}
        & \indshort & 0.0 & 0.0 & {NA} & 0 & 4 & 6713 & 0 & \indshort & 0.0 & 0.0 & {NA} & 0 & 4 & 6713 & 0\\
        \bottomrule
    \end{tabular}
    }
\end{table}

\para{Discussion}
As already mentioned in \cref{sec:experiments}, we only optimized the parameters regarding the $F_1$ score for predicting swarming events.
This has direct implications on the displayed metrics of \cref{tab:results_univariate,,tab:results_multivariate}, since any true anomaly that is not a swarm is reported as a false positive.
As we only have labels for swarming events, these tables are meant to show the differences in predictions when automatically optimizing models with very sparse (\wue: 8, \shw: 24 swarming windows) events and no specialization.

When comparing \cref{tab:results_univariate} predictions from both different training hives, there are a few remarkable differences: 
Overall, the classification results are better when training on \shw\ ($F_1 : [.09, .12]$) in contrast to training on \wue\ ($F_1: [.03,.12]$).
The main reason for that is the very high inclination of the classifiers towards never predicting a swarming event in \wue.
Some anomaly detectors even report no true positive for swarming events (Elliptic Envelope, One-Class SVM).
Even the metrics on the \jel\ test set decline significantly (\shw: $F_1: [.48, .69]$, \wue: $F_1: [.13, .45]$) for all detectors except the AE (\shw: $F_1: .73$, \wue: $F_1:.69$).

In both cases, the AE is the best swarm detector within the machine learning algorithms (highlighted in bold).
It also seems to be more robust regarding the origin of data, since the $F_1$ score (0.12) is the same for both training scenarios.

\algo\ is the best swarm detector regarding the metrics.
It does however miss more swarming events (4 vs. 24), some due to the windowing technique used, since in relies on the base temperature \SI{30}{\minute} pre-swarming.
The major contributing factor for the better metrics performance is the very low false positive rate.
This is to be expected, since it is only built for swarm detection and isn't drawn away towards other anomalies and thus inherently has a lower false positive rate.
For example, any temperature deviation below \SI{34.5}{\celsius} is completely ignored, but may in fact be an anomaly.

\subsection{Multivariate}
\cref{tab:results_multivariate} lists the classification metrics in the multivariate sensor setting for temperature sensors T$_6$, T$_7$, T$_8$ for the hives \shw\ and \wue.
\begin{table}[t!]
    \centering
    \caption{Overview of classification metrics and results in the multivariate setting.
    Results are only calculated by the true positives of swarms!
    The estimators are trained on \shw\, predicting \wue\ (and vice versa) and sensor T$_6$, T$_7$, T$_8$.
    Precision (P), Recall (R), and $F_1$ are reported, and set to 0 for no correct classification and $F_1$ set as NA.
    Corresponding true positives (TP), false positives (FP), false negatives (FN), and true negatives (TN) values are also reported.
    The overall metrics are calculated from the sum of the number of classifications.
    }
    \label{tab:results_multivariate}
    \begin{tabular}{@{}ll|S[round-mode=places,round-precision=3]S[round-mode=places,round-precision=3]S[round-mode=places,round-precision=3]|cccc@{}}
        \toprule
        Classifier & Beehive & {P[\%]} & {R[\%]} & {$F_1$[\%]} & TP & FP & TN & FN\\
        \midrule
        \multirow{3}{*}{\parbox{1cm}{Local\\Outlier\\Factor}} & \wue & 0 & 0 & {NA} & 0 & 55 & 4698 & 8\\
        & \shw & 0.005113221329437546 & 0.875 & 0.010167029774872912 & 21 & 4086 & 222 & 3\\
        & Overall & \bfseries 0.005045651129264777 & 0.65625 & 0.010014306151645207 & 21 & 4141 & 4920 & 11\\
        \cmidrule{2-9}
        & \ind$_W$ & 0 & 0 & {NA} & 0 & 5132 & 1585 & 0\\
        & \ind$_S$ & 0 & 0 & {NA} & 0 & 1055 & 6552 & 0\\
        
        \midrule
        
        \multirow{3}{*}{\parbox{1cm}{Elliptic\\Envelope}} & \wue & 0 & 0 & {NA} & 0 & 177 & 4576 & 8\\
        & \shw & 0.005536514632217242 & 0.875 & 0.01100340581608593 & 21 & 4063 & 578 & 3\\
        & Overall & \bfseries 0.00492842055855433 & 0.65625 & 0.009783368273934312 & 21 & 4240 & 5154 & 11\\
        \cmidrule{2-9}
        & \ind$_W$ & 0 & 0 & {NA} & 0 & 3955 & 3762 & 0\\
        & \ind$_S$ & 0 & 0 & {NA} & 0 & 2842 & 3875 & 0\\
        
        \midrule
        
        \multirow{3}{*}{\parbox{1cm}{Isolation\\Forest}} & \wue & 0.0012461059190031153 & 0.5 & 0.0024860161591050344 & 4 & 3206 & 1547 & 4\\
        & \shw & 0.0054375970999482135 & 0.875 & 0.010808028821410192 & 21 & 3841 & 800 & 3\\
        & Overall & 0.0035350678733031674 & 0.78125 & 0.007038288288288288 & 25 & 7047 & 2347 & 7\\
        \cmidrule{2-9}
        & \ind$_W$ & 0 & 0 & {NA} & 0 & 6694 & 23 & 0\\
        & \ind$_S$ & 0 & 0 & {NA} & 0 & 6387 & 330 & 0\\
        
        \midrule
        
        \multirow{3}{*}{\parbox{1cm}{One-Class\\SVM}} & \wue & 0.0012461059190031153 & 0.5 & 0.0024860161591050344 & 4 & 3206 & 1547 & 4\\
        & \shw & 0.0054375970999482135 & 0.875 & 0.010808028821410192 & 21 & 3841 & 800 & 3\\
        & Overall & 0.0035350678733031674 & 0.78125 & 0.007038288288288288 & 25 & 7047 & 2347 & 7\\
        \cmidrule{2-9}
        & \ind$_W$ & 0 & 0 & {NA} & 0 & 6694 & 23 & 0\\
        & \ind$_S$ & 0 & 0 & {NA} & 0 & 6387 & 330 & 0\\
        
        \midrule
        
        \multirow{3}{*}{AE} & \wue & 0.0007733952049497294 & 0.125 & 0.0015372790161414297 & 1 & 1292 & 3461 & 7\\
        & \shw & 0.007337526205450734 & 0.875 & 0.014553014553014554 & 21 & 2841 & 1800 & 3\\
        & Overall & \bfseries 0.00529482551143201 & 0.6875 & \bfseries 0.010508717458801052 & 22 & 4133 & 5261 & 10\\
        \cmidrule{2-9}
        & \ind$_W$ & 0 & 0 & {NA} & 0 & 6683 & 43 & 0\\
        & \ind$_S$ & 0 & 0 & {NA} & 0 & 5534 & 1183 & 0\\
        \bottomrule
    \end{tabular}
\end{table}
Calculation of the metrics is done in the same manner as in the univariate setting.
The table lists only the multivariate datasets.
The classification metrics of \wue\ are reported when training on \shw\, and vice versa.
Furthermore the lines with \ind$_X$ show the reports when training on hive $X$ and predicting \ind.

\begin{figure}[H]
    \centering
    \includegraphics[width=.8\textwidth]{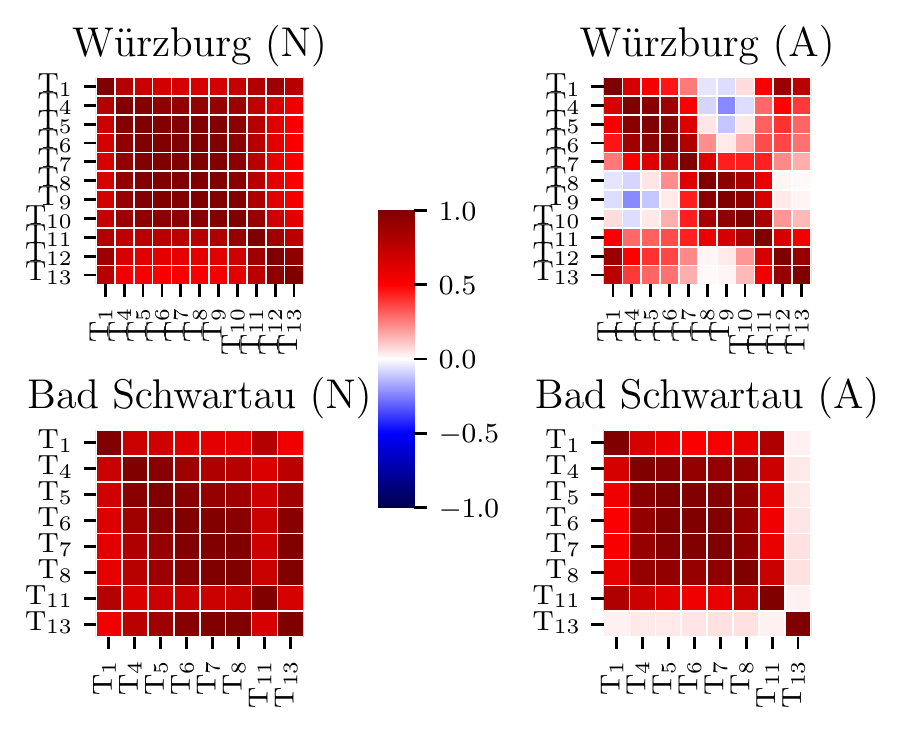}
    \caption{Sensor correlations. All figures display the Pearson correlation between temperature sensors within a given beehive. (N) stands for the dataset containing normal behavior and (A) for the dataset with anomalous behavior.~\cite{davidson2020anomaly}}
    \label{fig:correlation_anomalies}
\end{figure}

\para{Discussion}
In the multivariate setting, the AE is also the best option for detecting anomalies.
Still \cref{tab:results_multivariate} shows, that all metrics drop in contrast to the univariate, single temperature sensor setting.
The reason for that is the much higher false positive rate, which means, that more non-swarms are confused with swarms.
This means the additional measurements introduce more noise as would be necessary for predicting swarms.
As shown in \cref{fig:correlation_anomalies}, adjacent sensors correlate strongly within the normal data, thus they bear no additional information during training, but weaker so within the anomaly set.
Only including new sources of information (like the scale) would help in the multivariate sensor setting, as shown in \cref{fig:colony_behavior_triple_swarm,,fig:colony_behavior_swarm_not_detected}.

\subsection{Methodology}
In \cref{sec:dataset} we described our empirically founded, but manual approach of splitting data into anomalous behavior and normal sensor data.
However, this data splitting method is ambiguous and highly susceptible to missing days in the corresponding dataset, i.e. missing anomalies and therefore mislabeling specific days.
A general, rule-based approach of splitting anomalous and normal data, i.e. all windows with sensor values drifting for more than two standard deviations, doesn't work, since it removes most swarming events from the test set.
A clearer split of training and testing data can only be ensured by very thorough labeling of the sensor values, which has to be done on different sensors independently.

In this work we evaluated predictions in an automated manner by using a random search for the best parameter settings (\cref{tab:appendix_hyperparameters}) using only labelled information of swarming events.
In previous work \cite{davidson2020anomaly} we selected the parameter $\alpha$ for the AE for detecting anomalies manually.
This is a first step towards the automation of the anomaly detectors, but still has the problem of only being optimized for one anomaly class and still results in false positives for swarming events, but true positives for other anomalies.

Summarizing the results, the AE is the best all purpose swarm detector within the machine learning algorithms.
It is out-performed to \algo\ for swarming detection, but it is also capable of predicting other anomalies without the knowledge of special rules.

\section{Analysis}
\label{sec:discussion}
In this section we analyze the found anomalies and will outline different types of anomalies reported by the AE.
We used this model exemplary to show interesting observations from the predictions, not only focussing on swarms, but also the aforementioned false positives, as well as true positives for other anomalies, hidden from the above discussion.

\begin{figure}[b!]
        \centering
        \includegraphics[width=0.7\textwidth]{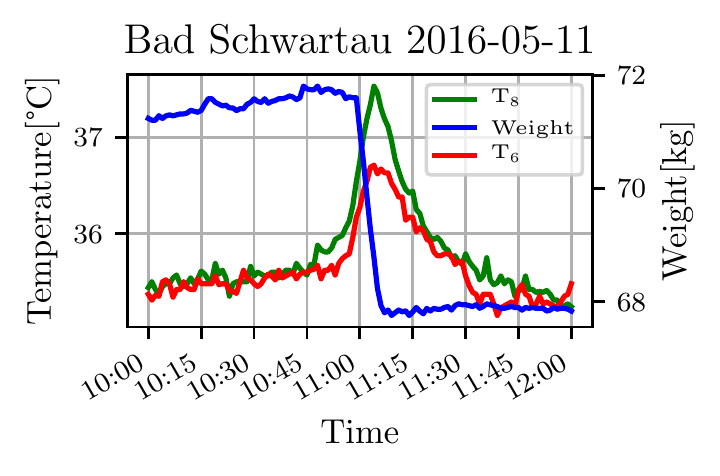}
        \caption{(Prototypical) Swarm as indicated by T$_6$ and T$_8$, detected by \algo\ and AE.\cite{davidson2020anomaly}}
        \label{fig:colony_behavior_swarm_real_detected}
\end{figure}

\para{Swarming events}
All swarming events predicted with temperature sensors T$_6$ and T$_m$ by the AE and \algo\ can be found in \cref{tab:results_predictions}.
\begin{table}[t!]
    \centering
    \caption{Detected Anomalies. The first column shows the shortened name of the used test (anomaly) set (\shwshort{} is \shw, \wueshort{} is \wue, \jelshort{} is \jel, \indshort{} is \ind).
    (S) signifies that the set contains swarms while (O) stands for other anomalies.
    The next column displays the date of the event, and --- where suitable --- a reference to figures in the text.
    The last two columns indicate whether \algo\ or our method (AE) detected the anomaly.
    Predictions on \hobos-hives are based on sensor T$_6$, on T$_m$ for \webee.
    We used the \shw\ trained model to predict the swarms in any other beehive, except for \shw\ itself.~\cite{davidson2020anomaly}
    }

    \begin{tabular}{@{}llcc@{}}
        \toprule
        \multirow{2}{*}{Dataset} & \multirow{2}{*}{Timestamp} & \multicolumn{2}{c}{Detected}\\
        \cmidrule{3-4}
        && \algo & AE\\
        \midrule
        \multirow{7}{*}{\shwshort~(S)} & 2016-05-11 11:05\textsuperscript{\ref{fig:colony_behavior_swarm_real_detected}} & $\checkmark$ & $\checkmark$\\
         & 2016-05-22 07:30 & $\checkmark$ & $\checkmark$\\
         & 2017-06-06 15:02 & $\checkmark$ & $\checkmark$\\
         & 2019-05-13 09:30$^\star$ & $\checkmark$ & $\checkmark$\\
         & 2019-05-21 09:15$^\star$ & $\checkmark$ & $\checkmark$\\
         & 2019-05-25 12:00$^\star$ & $\checkmark$ & $\checkmark$\\
         \cmidrule{2-4}
        \shwshort~(O) & 2016-08-03 17:24 & $\checkmark$ & $\checkmark$\\
        \midrule
        \multirow{2}{*}{\wueshort~(S)} & 2019-05-01 09:15\textsuperscript{\ref{fig:colony_behavior_swarm_not_detected}} & & $\checkmark$\\
        & 2019-05-10 11:15\textsuperscript{\ref{fig:colony_behavior_triple_swarm}} & $\checkmark$ & $\checkmark$\\
        \cmidrule{2-4}
        \wueshort~(O) & 2019-04-17 16:22\textsuperscript{\ref{fig:colony_behavior_swarm_detected}} & $\checkmark$ & $\checkmark$\\
        \bottomrule
    \end{tabular}
    \hspace{1cm}
    \begin{tabular}{@{}llcc@{}}
        \toprule
        \multirow{2}{*}{Dataset} & \multirow{2}{*}{Timestamp} & \multicolumn{2}{c}{Detected}\\
        \cmidrule{3-4}
        && \algo & AE\\
        \midrule
        \multirow{9}{*}{\jelshort~(S)} & 2015-05-06 18:02$^\star$ & $\checkmark$ & $\checkmark$\\
         & 2016-06-02 13:48$^\star$ & $\checkmark$ & $\checkmark$\\
         & 2016-05-30 10:03$^\star$ & $\checkmark$ & $\checkmark$\\
         & 2016-06-16 15:50$^\star$ & $\checkmark$ & $\checkmark$\\
         & 2016-06-01 13:20$^\star$ & $\checkmark$ & $\checkmark$\\
         & 2016-06-03 09:11$^\star$ & $\checkmark$ & $\checkmark$\\
         & 2016-06-13 03:30 & $\checkmark$ & $\checkmark$\\
         & 2016-06-16 10:52$^\star$ & $\checkmark$ & $\checkmark$\\
         & 2016-06-13 13:32$^\star$ & $\checkmark$ & $\checkmark$\\
         \midrule
        \multirow{2}{*}{\indshort~(O)} & 2019-07-26 08:10 & $\checkmark$ & $\checkmark$\\
         & 2019-08-31 17:08\textsuperscript{\ref{fig:colony_behavior_varroa}} & $\checkmark$ & \\
        \bottomrule
    \end{tabular}
    \label{tab:results_predictions}
\end{table}
Events observed by apiarists on site are marked with $^\star$.
This table lists all swarm like events detected by \algo, as well as additionally missed swarming events.
In other words, we used \algo\ to verify the results of the AE and vise versa, as described in~\cite{davidson2020anomaly}.
\cref{fig:colony_behavior_swarm_real_detected} shows a sensor data plot for a prototypical swarm for the hive in \shw.
Swarming events can be found within the table as \textit{location (S)}, whereas other anomalies are denoted with \textit{location (O)}.
A more detailed view of the findings regarding swarming events is given in \cite{davidson2020anomaly}.

\begin{figure}[b!]
    \centering
    \begin{subfigure}[c]{0.475\textwidth}    
        \includegraphics[width=\textwidth]{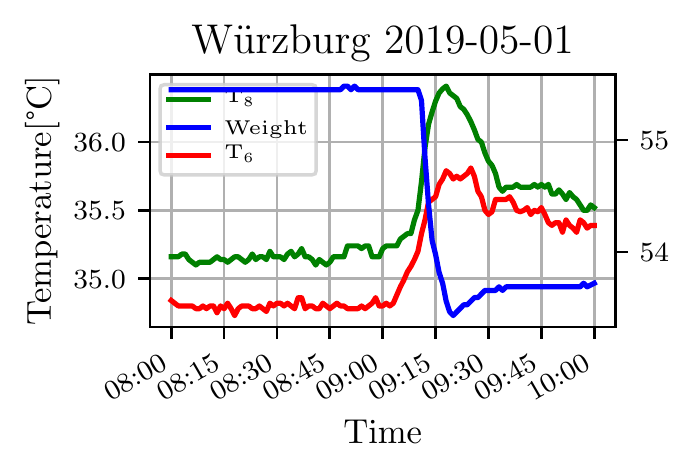}
        \caption{Swarm detected with T$_8$, but not with T$_6$ (\algo).
        Anomaly in both for AE. Swarm anomaly within the weight.\cite{davidson2020anomaly}}
        \label{fig:colony_behavior_swarm_not_detected}
    \end{subfigure}
    ~
    \begin{subfigure}[c]{0.455\textwidth}
        \includegraphics[width=\textwidth]{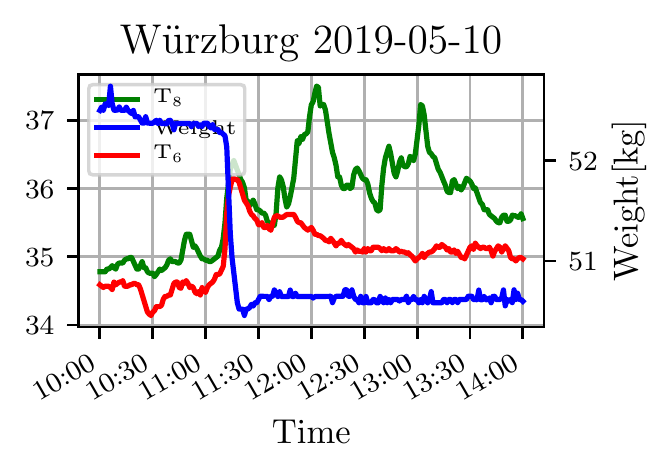}
        \caption{Swarm anomaly indicated by both T$_6$ and T$_8$, but additional swarms in T$_8$. Swarm anomaly within the weight.\cite{davidson2020anomaly}}
        \label{fig:colony_behavior_triple_swarm}
    \end{subfigure}
    ~
    \begin{subfigure}[t]{0.5\textwidth}
        \includegraphics[width=\textwidth]{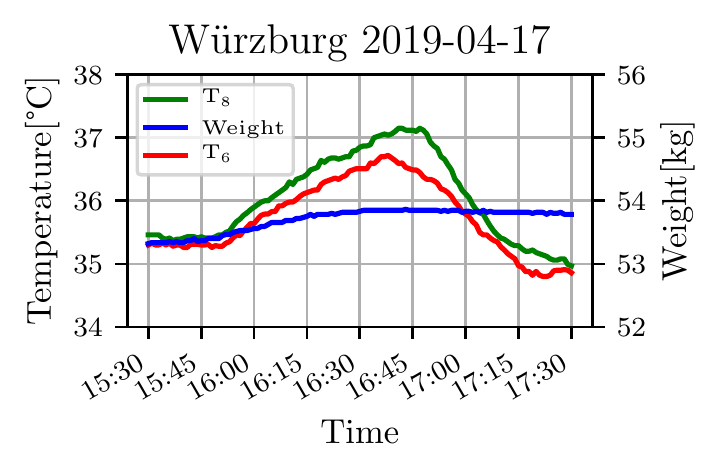}
        \caption{Swarm-like anomaly in sensors T$_6$ and T$_8$, but not within the measured weight.\cite{davidson2020anomaly}}
        \label{fig:colony_behavior_swarm_detected}
    \end{subfigure}
    \caption{Special cases of swarming events.
    (a) shows a swarm only detected with one temperature sensor, but not the other (\algo).
    (b) shows a swarming event followed by subsequent swam-like temperature curves in T$_8$.
    (c) shows a swarm-like anomaly in the temperature sensors, but not in the scale.
    }
    \label{fig:colony_behavior_swarms}
\end{figure}

\begin{figure}[b!]
    \centering
    \begin{subfigure}[t]{0.475\textwidth}    
        \includegraphics[width=\textwidth]{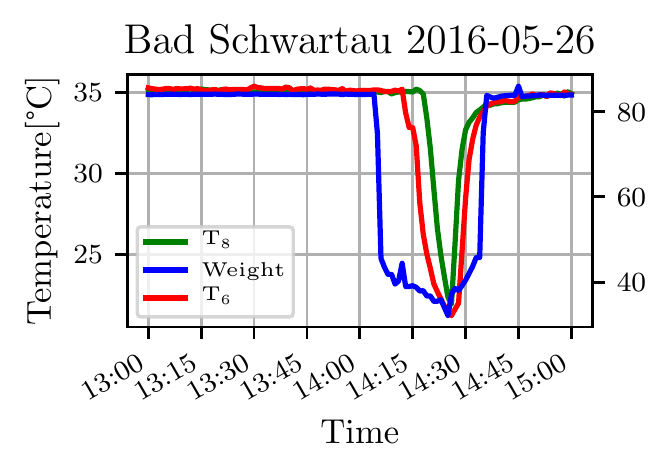}
        \caption{External interference of an opened apiary.
        The influx of outer air leads to the temperature drop.\cite{davidson2020anomaly}}
        \label{fig:colony_behavior_anomaly_open_hive}
    \end{subfigure}
    ~
    \begin{subfigure}[t]{0.475\textwidth}
        \includegraphics[width=\textwidth]{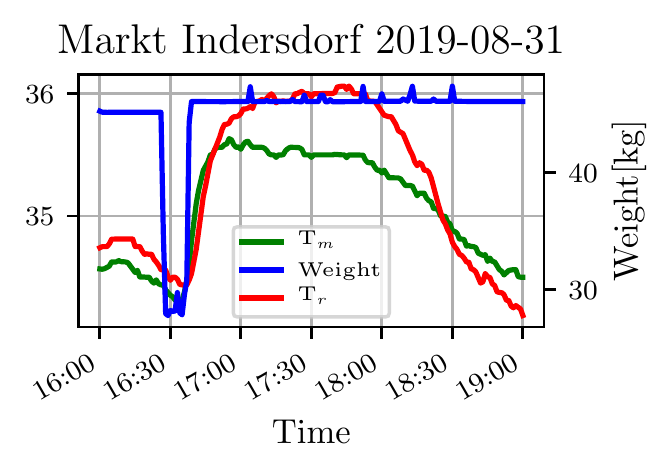}
        \caption{External interference by a possible varroa treatment.
        The beehive was opened, weight added, leading to the excitement of bees with a temperature increase. In contrast to our AE with T$_m$, \algo\ detected a swarm with T$_r$ and T$_m$.\cite{davidson2020anomaly}}
        \label{fig:colony_behavior_varroa}
    \end{subfigure}
    ~
    \begin{subfigure}[t]{0.5\textwidth}
        \includegraphics[width=\textwidth]{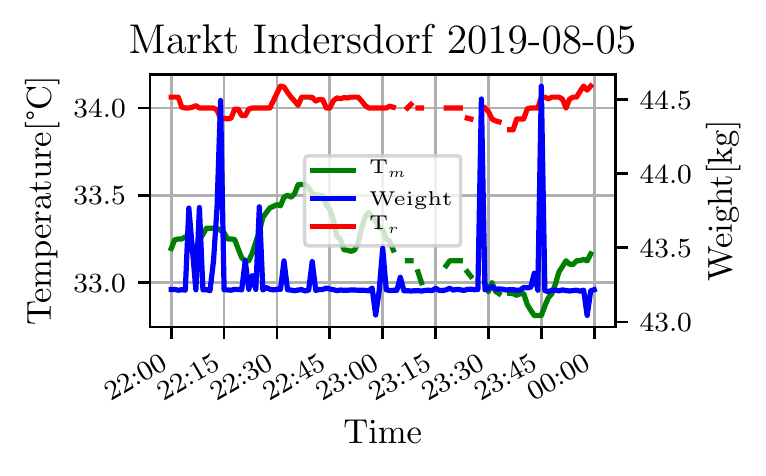}
        \caption{Sensor anomaly with missing values in T$_r$ and T$_m$, but not in the measured weights.\cite{davidson2020anomaly}}
        \label{fig:colony_behavior_sensor_defect}
    \end{subfigure}
    \caption{External interference anomalies.
    (a) shows an opened hive with no modifications, whereas (b) is opened for treatment with a substance added.
    (c) shows missing sensor values.
    }
    \label{fig:colony_behavior_interference}
\end{figure}

\para{Other anomalies}
\cref{fig:colony_behavior_swarms} depicts anomalies easily confused with swarms in at least one sensor.
\cref{fig:colony_behavior_interference} on the other hand, show anomalies categorized as external interference.
They all display the same sensors, two temperature sensors (\hobos: T$_6$, T$_8$; \webee: T$_r$, T$_m$) and the weight on the scale.
An exemplary plot of a training sample can be seen in \cref{fig:colony_behavior_normal}.
Sensors in \cref{fig:colony_behavior_swarm_real_detected} show the expected behavior for a swarming event, as already stated in \cref{sec:experiments}.

Detecting swarms only in traces of temperature data, also has its drawbacks, as \cref{fig:colony_behavior_swarm_detected} shows, since the values of the weight sensors tend to describe normal behavior, whereas the temperature sensors follow the expected inverted parabola.

Similar implications can be seen in \cref{fig:colony_behavior_triple_swarm}, as a slice of the window actually contains a swarm, shown by all three sensors, whereas a later slice only indicates a swarm temperature-wise.

\cref{fig:colony_behavior_swarm_not_detected} shows a swarming event, which \algo\ only detects in T$_8$, but not T$_6$, since it is not covered by the defined rules for swarms.
The AE on the other hand is capable of detecting this swarm in both temperature sensors.

\cref{fig:colony_behavior_anomaly_open_hive} depicts the sensor traces of an opened apiary, which becomes obvious in the fast and strong drop in weight, and with varying delay in time, in the temperature sensors.
This is due to the influx of ambient air, cooling the temperature within the beehive.
As soon as the hive is closed the expected values, the same as before opening, are reported again.

The beehive must sometimes be opened for treatment purposes.
An example of a varroa treatment with a substance (i.e. formic acid) is displayed in \cref{fig:colony_behavior_varroa}.
The resulting additional weight after closing the hive in visible in the weight sensor.
\algo\ confuses this as a swarm in both temperature sensors, whereas the AE only reports a swarm for T$_r$.
T$_m$ only fluctuates within one standard deviation of training data, which can be captured by the feature space of the AE.

Aforementioned anomalies are only a subset of reported anomalies, since the AE detects a lot more.
Some of them are not as easily classified, but normally are temperature values far lower than \SI{30}{\celsius}.
Even sensor anomalies are detected by AE, as can be seen in \cref{fig:colony_behavior_sensor_defect}.




\section{Conclusion/Future Work}
\label{sec:conclusion}
\noindent
In this work we evaluated the use of machine learning models for anomaly detection in beehives.
We compared the models Elliptic Envelope, Isolation Forests, Local Outlier Factor, One-Class SVMs, and recurrent autoencoders quantitatively for swarm detection.
The results show that the AE is the best multi-purpose anomaly detector in comparison.
It is able to detect swarms with high accuracy even by only optimizing the decision threshold with very sparse swarm instances.
Within the multivariate temperature sensor setting we found, that combining there sensors incurs more noise than information, and still needs further experiments and evaluation.
Especially combining different sensor types, i.e. temperature and weight, seems to be more promising.
Multiple aspects of anomaly detection in beehives require more work in the future:





\para{Evaluation of deep generative models}
Other types of deep neural networks will have to be explored in future work.
For example, generative models like variational autoencoders or generative adversarial networks show particular promise, since they have two advantages:
\begin{enumerate*}
    \item[A)] anomalies may exist within the training set, and
    \item[B)] they allow for probability-based classification instead of relying solely on the reconstruction error~\cite{an2015variational}.
\end{enumerate*}


\para{Dataset generation}
Machine learning models require data to correctly learn their task.
The amount of beehive data available is limited, especially when considering data with labeled anomalies like swarming.
To this end, we hope to improve data availability in the project \webee, where sensor-equipped apiaries are distributed mostly across Germany, allowing us to collect a large dataset of beehive data.
Any events or anomalies can be marked by apiarists participating in \webee, providing us with more valuable labeled data.
Predictive alert-systems can be implemented to warn beekeepers in case of anomalies.
The beekeepers may provide feedback for the warnings, which allows further improvements in prediction quality.


\para{Winter period}
During winter, bees enter a passive state where their behavior changes significantly in comparison to summer time~\cite{zacepins2015challenges}.
To learn normal behavior of bees for their active summer time, we excluded data from October through March for all datasets (cf. \cref{sec:dataset}).
Detecting anomalies during winter can also be of interest but this remains future work.


\section*{Acknowledgements}
\noindent
This research was conducted in the we4bee project sponsored by the Audi Environmental Foundation.

\section*{Appendix}
\subsection*{Sensor correlations}

\subsection*{Hyperparameters}
\begin{table}[H]
    \centering
    \caption{
    Optimized parameters and their ranges for the anomaly detectors within the random search.
    $\mathcal{U}_x$ describes an uniform distribution with $[0, x)$, whereas $\mathcal{I}_{a,b}$ represents a random integer distribution with $[a, b]$.
    $\mathcal{LU}_{a, b}$ is a log uniform distribution with parameters $a, b$.
    }
    \label{tab:appendix_hyperparameters}
    \begin{tabular}{@{}l|cc@{}}
    \toprule
    Classifier & Hyperparameter & Range\\
    \midrule
    \multirow{5}{*}{\parbox{1cm}{Local\\Outlier\\Factor}} & \textit{n neighbors} & $\mathcal{I}_{1, 100}$\\
     & \textit{algorithm} & ball tree, kd tree\\
     & \textit{leaf size} & $\mathcal{I}_{1, 150}$\\
     & \textit{contamination} & $\mathcal{U}_{0.5}$\\
     & \textit{metric} & chebyshev, cityblock, euclidean, infinity, l1, l2, manhattan, minkowski\\
    \midrule
    \multirow{3}{*}{\parbox{1cm}{Elliptic\\Envelope}} & \textit{assume centered} & True, False\\
     & \textit{support fraction} & $\mathcal{U}_1$\\
     & \textit{contamination} & $\mathcal{U}_{0.5}$\\
    \midrule
    \multirow{6}{*}{\parbox{1cm}{Isolation\\Forest}} & \textit{n estimators} & $\mathcal{I}_{10, 100}$\\
    & \textit{max samples} & auto\\
    & \textit{contamination} & $\mathcal{U}_{0.5}$\\
    & \textit{max features} & $\mathcal{U}_1$\\
    & \textit{bootstrap} & True, False\\
    \midrule
    \multirow{4}{*}{\parbox{1cm}{One-\\Class\\SVM}} & \textit{kernel} & linear, poly(degree=3), rbf(coef0=0), sigmoid\\
     & \textit{shrinking} & True, False\\
     & $\gamma$ & $\mathcal{LU}_{0.0001, 1}$\\
     & $\nu$ & $\mathcal{LU}_{0.0001, 1}$\\
    \midrule
    \multirow{2}{*}{\parbox{1cm}{AE}} & \textit{hidden size} & $\mathcal{I}_{2, 64}$\\
     & \textit{layers} & $\mathcal{I}_{1, 4}$\\
    \bottomrule
    \end{tabular}
\end{table}

\clearpage
\bibliographystyle{splncs04}
\bibliography{bibliography}

\begin{thebibliography}{10}
\providecommand{\url}[1]{\texttt{#1}}
\providecommand{\urlprefix}{URL }
\providecommand{\doi}[1]{https://doi.org/#1}

\bibitem{an2015variational}
An, J., Cho, S.: Variational autoencoder based anomaly detection using
  reconstruction probability. Special Lecture on IE  \textbf{2}(1) (2015)

\bibitem{bergstra2012random}
Bergstra, J., Bengio, Y.: Random search for hyper-parameter optimization. JMLR
  \textbf{13},  281--305 (2012)

\bibitem{breunig2000lof}
Breunig, M.M., Kriegel, H.P., Ng, R.T., Sander, J.: Lof: identifying
  density-based local outliers. In: Proceedings of the 2000 ACM SIGMOD
  international conference on Management of data. pp. 93--104 (2000)

\bibitem{chalapathy2019deepad}
Chalapathy, R., Chawla, S.: Deep learning for anomaly detection: {A} survey.
  CoRR  \textbf{abs/1901.03407} (2019), \url{http://arxiv.org/abs/1901.03407}

\bibitem{cortes1995svm}
Cortes, C., Vapnik, V.: Support-vector networks. Machine Learning
  \textbf{20}(3),  273--297 (Sep 1995). \doi{10.1007/BF00994018},
  \url{https://doi.org/10.1007/BF00994018}

\bibitem{davidson2020anomaly}
Davidson, P., Steininger, M., Lautenschlager, F., Kobs, K., Krause, A., Hotho,
  A.: Anomaly detection in beehives using deep recurrent autoencoders. In:
  Proceedings of the 9th International Conference on Sensor Networks
  (SENSORNETS 2020). pp. 142--149. No.~9, SCITEPRESS – Science and Technology
  Publications, Lda. (2020)

\bibitem{fell1977seasonal}
Fell, R.D., Ambrose, J.T., Burgett, D.M., De~Jong, D., Morse, R.A., Seeley,
  T.D.: The seasonal cycle of swarming in honeybees. Journal of Apicultural
  Research  \textbf{16}(4),  170--173 (1977)

\bibitem{ferrari2008monitoring}
Ferrari, S., Silva, M., Guarino, M., Berckmans, D.: Monitoring of swarming
  sounds in bee hives for early detection of the swarming period. Computers and
  electronics in agriculture  \textbf{64}(1),  72--77 (2008)

\bibitem{filonov2016multivariate}
Filonov, P., Lavrentyev, A., Vorontsov, A.: Multivariate industrial time series
  with cyber-attack simulation: Fault detection using an {LSTM}-based
  predictive data model. NIPS Time Series Workshop 2016  (2016)

\bibitem{kingma2014adam}
Kingma, D.P., Ba, J.: Adam: A method for stochastic optimization. arXiv
  preprint arXiv:1412.6980  (2014)

\bibitem{kridi2014predictive}
Kridi, D.S., Carvalho, C.G.N.d., Gomes, D.G.: A predictive algorithm for
  mitigate swarming bees through proactive monitoring via wireless sensor
  networks. In: Proceedings of the 11th ACM symposium on PE-WASUN. pp. 41--47.
  ACM (2014)

\bibitem{li2003improving}
Li, K.L., Huang, H.K., Tian, S.F., Xu, W.: Improving one-class svm for anomaly
  detection. In: Proceedings of the 2003 International Conference on Machine
  Learning and Cybernetics (IEEE Cat. No. 03EX693). vol.~5, pp. 3077--3081.
  IEEE (2003)

\bibitem{liu2008isolation}
Liu, F.T., Ting, K.M., Zhou, Z.H.: Isolation forest. In: 2008 Eighth IEEE
  International Conference on Data Mining. pp. 413--422. IEEE (2008)

\bibitem{liu2012isolation}
Liu, F.T., Ting, K.M., Zhou, Z.H.: Isolation-based anomaly detection. ACM
  Transactions on Knowledge Discovery from Data (TKDD)  \textbf{6}(1),  1--39
  (2012)

\bibitem{malhotra2016multi}
Malhotra, P., Tv, V., Ramakrishnan, A., Anand, G., Vig, L., Agarwal, P.,
  Shroff, G.: Multi-sensor prognostics using an unsupervised health index based
  on lstm encoder-decoder. 1st SIGKDD Workshop on ML for PHM  (08 2016)

\bibitem{scikit-learn2011sciki-learn}
Pedregosa, F., Varoquaux, G., Gramfort, A., Michel, V., Thirion, B., Grisel,
  O., Blondel, M., Prettenhofer, P., Weiss, R., Dubourg, V., Vanderplas, J.,
  Passos, A., Cournapeau, D., Brucher, M., Perrot, M., Duchesnay, E.:
  Scikit-learn: Machine learning in {P}ython. Journal of Machine Learning
  Research  \textbf{12},  2825--2830 (2011)

\bibitem{rousseeuw1984least}
Rousseeuw, P.J.: Least median of squares regression. Journal of the American
  statistical association  \textbf{79}(388),  871--880 (1984)

\bibitem{rousseeuw1999fast}
Rousseeuw, P.J., Driessen, K.V.: A fast algorithm for the minimum covariance
  determinant estimator. Technometrics  \textbf{41}(3),  212--223 (1999)

\bibitem{ryan1998intrusion}
Ryan, J., Lin, M.J., Miikkulainen, R.: Intrusion detection with neural
  networks. In: Advances in neural information processing systems. pp. 943--949
  (1998)

\bibitem{scholkopf2001estimating}
Sch{\"o}lkopf, B., Platt, J.C., Shawe-Taylor, J., Smola, A.J., Williamson,
  R.C.: Estimating the support of a high-dimensional distribution. Neural
  computation  \textbf{13}(7),  1443--1471 (2001)

\bibitem{shipmon2017time}
Shipmon, D.T., Gurevitch, J.M., Piselli, P.M., Edwards, S.T.: Time series
  anomaly detection; detection of anomalous drops with limited features and
  sparse examples in noisy highly periodic data. arXiv preprint
  arXiv:1708.03665  (2017)

\bibitem{winston1980swarming}
Winston, M.: Swarming, afterswarming, and reproductive rate of unmanaged
  honeybee colonies (apis mellifera). Insectes Sociaux  \textbf{27}(4),
  391--398 (1980)

\bibitem{zacepins2015challenges}
Zacepins, A., Brusbardis, V., Meitalovs, J., Stalidzans, E.: Challenges in the
  development of precision beekeeping. Biosystems Engineering  \textbf{130},
  60--71 (2015)

\bibitem{zacepins2016remote}
Zacepins, A., Kviesis, A., Stalidzans, E., Liepniece, M., Meitalovs, J.: Remote
  detection of the swarming of honey bee colonies by single-point temperature
  monitoring. Biosystems engineering  \textbf{148},  76--80 (2016)

\bibitem{zhou2017anomaly}
Zhou, C., Paffenroth, R.C.: Anomaly detection with robust deep autoencoders.
  In: Proceedings of the 23rd ACM SIGKDD. pp. 665--674. ACM (2017)

\bibitem{zhu2019increase}
Zhu, X., Wen, X., Zhou, S., Xu, X., Zhou, L., Zhou, B.: The temperature
  increase at one position in the colony can predict honey bee swarming (apis
  cerana). Journal of Apicultural Research  \textbf{58}(4),  489--491 (2019)

\end{thebibliography}


\end{document}